\documentclass[sigconf,nonacm]{acmart}
\setcopyright{none}
\renewcommand{\shortauthors}{Wang and Lu}
\usepackage{amsmath,mathtools}
\usepackage{booktabs,multirow,array,makecell}
\usepackage{graphicx,enumitem}
\usepackage[table]{xcolor}
\usepackage{balance,capt-of,multicol}
\usepackage{microtype}
\usepackage{placeins}
\usepackage{stfloats}
\usepackage{etoolbox}

\usepackage{pifont}
\definecolor{rankbest}{HTML}{FFF2CC}
\definecolor{ranksecond}{HTML}{F4B183}
\definecolor{rankthird}{HTML}{DDEBF7}
\definecolor{oraclegray}{HTML}{E7E6E6}

\newcommand{\bestcell}[1]{\cellcolor{rankbest}#1}
\newcommand{\secondcell}[1]{\cellcolor{ranksecond}#1}
\newcommand{\thirdcell}[1]{\cellcolor{rankthird}#1}
\newcommand{\cmark}{\textcolor{green!55!black}{\ding{51}}}
\newcommand{\xmark}{\textcolor{red!75!black}{$\times$}}
\definecolor{bestorange}{HTML}{F6C4A8}
\definecolor{secondyellow}{HTML}{F5E4B2}
\definecolor{thirdblue}{HTML}{DDEAF3}
\definecolor{faintgray}{HTML}{ECECEC}

\newcommand{\Usct}{u^{\mathrm{SCT}}}
\newcommand{\Rn}{R_{\mathrm{naive}}}
\newcommand{\ssrc}{\mathrm{src}}
\newcommand{\starc}{\mathrm{tar}}

\graphicspath{{figure/}}
\title{PreciseEdit: Strong Yet Stable Semantic Transport for One-Step Image Editing}

\author{Jingyi Wang}
\email{jingyiwang@hainanu.edu.cn}
\affiliation{
  \institution{Hainan University}
  \department{School of Computer Science and Technology}
  \city{Haikou}
  \country{China}
}

\author{Xiao Lu}
\email{lu\_x@shu.edu.cn}
\affiliation{
  \institution{Shanghai University}
  \department{School of Mechatronic Engineering and Automation}
  \city{Shanghai}
  \country{China}
}

\begin{abstract}
Recent one-step text-to-image (T2I) models enable efficient image synthesis, opening up new possibilities for interactive Web-based image editing. However, existing one-step editing methods often construct a global transport field through simple vector operations between conditional predictions. While such strong transport can effectively drive the image toward the target semantics, it may be released along unreliable directions or extend into non-edited regions, leading to structural distortions within the editing region and background drift elsewhere. More importantly, stabilizing the editing process through global smoothing or uniformly weakening the transport field often makes large semantic changes overly conservative. In this work, we revisit one-step image editing from the perspective of the tension between semantic strength and transport stability, and identify a key insight: strong semantic transport is not inherently unstable. What matters is whether it is applied along reliable directions and at reliable spatial locations. Motivated by this insight, we introduce PreciseEdit, a model-agnostic, training-free and inversion-free one-step editing framework that strengthens target-semantic alignment while preserving background fidelity and structural consistency. We further propose a dual-noise-scale role decomposition strategy: a higher-noise scale produces the strong target-directed transport used for actual editing, whereas a lower-noise scale jointly serves as a probe for directional agreement and spatial structure. Extensive experiments on PIE-Bench demonstrate that PreciseEdit achieves state-of-the-art performance and effectiveness on various editing tasks.
\end{abstract}
\makeatletter
\def\@authorfont{\fontsize{10.8}{11.5}\selectfont}
\def\@affiliationfont{\fontsize{8.8}{9.9}\selectfont}
\makeatother
\begin{document}
\nocite{avrahami2022blended,cao2023masactrl,couairon2023diffedit,dao2024swiftbrushv2,deutch2024turboedit,dong2023prompttuning,fu2026layeredit,geng2025meanflow,gong2025instantedit,han2024proxedit,hertz2023prompt2prompt,ho2020ddpm,hu2026whereedit,ju2024pnpinversion,kawar2023imagic,kulikov2025flowedit,liu2024instaflow,lu2026chordedit,mao2024magedit,meng2022sdedit,mokady2023nulltext,nguyen2024swiftbrush,nguyen2025swiftedit,peng2026facm,radford2021clip,rombach2022ldm,sauer2024add,simsar2025lime,song2021ddim,song2023consistency,soni2025locatedit,starodubcev2024icd,tang2024locinv,tumanyan2023pnp,wallace2023edict,xu2024infedit,yin2024dmd2,yin2024dmd}
\maketitle
\pagestyle{plain}
\begin{figure}[t]
 \centering\includegraphics[width=0.965\columnwidth]{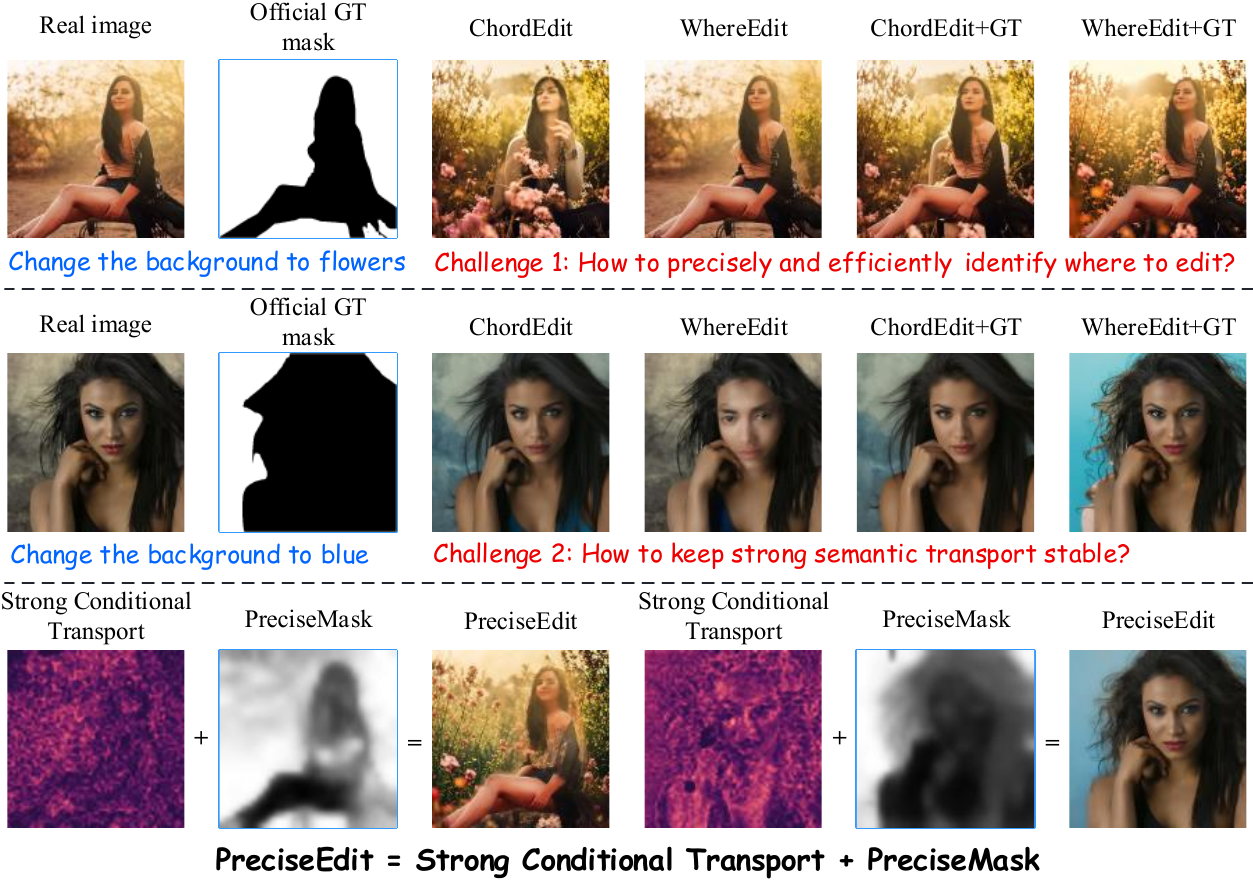}
 \caption{Motivation of PreciseEdit. The first row highlights the importance of reliable spatial localization: unconstrained transport may alter non-target content, while inaccurate localization may miss the intended edit region, raising our first challenge. The second row illustrates the strength--stability dilemma in one-step editing: ChordEdit~\cite{lu2026chordedit} remains conservative even with accurate spatial guidance, whereas the stronger transport of WhereEdit~\cite{hu2026whereedit} achieves larger semantic changes but may affect non-target regions, raising our second challenge. These observations motivate our selective-release principle for strong semantic transport.}
 \label{fig:motivation}
\end{figure}
\section{Introduction}
Modern Web platforms support visual creation and editing across social media, e-commerce, and online design tools, where users revise uploaded images through natural-language instructions. Recent one-step text-to-image (T2I) models, including SD-Turbo~\cite{sauer2024add}, SwiftBrush-v2~\cite{dao2024swiftbrushv2}, and InstaFlow~\cite{liu2024instaflow}, substantially reduce generation latency through one-step inference, making image creation increasingly practical. These advances motivate real-time text-guided editing for Web-based multimedia content creation, where users expect responsive feedback and precise control over changes. A key challenge is to achieve strong target-semantic changes with low latency while preserving non-edited content.

However, efficient one-step editing that enables strong semantic transformations while preserving high fidelity remains challenging. Some fast editors rely on trained inversion networks or explicit inversion~\cite{gong2025instantedit,nguyen2025swiftedit}, while training-free and inversion-free approaches construct editing directions directly from conditional predictions~\cite{kulikov2025flowedit,xu2024infedit}. Yet, compressing semantic transformation into a single update creates a pronounced strength--stability dilemma. ChordEdit~\cite{lu2026chordedit} stabilizes editing through temporal smoothing and a low-energy field but can under-realize large semantic changes, whereas the concurrent work WhereEdit~\cite{hu2026whereedit} enables stronger spatially constrained editing but incurs extra U-Net evaluation costs while its localization remains relatively coarse.

Despite their different formulations, these approaches expose two prominent drawbacks. First, globally weakening transport improves preservation but limits semantic transformations. Second, stronger transport improves target-semantic alignment but can cause background drift or structural distortion when applied unreliably. These issues naturally lead us to reconsider the current paradigm and pose a critical question: \textbf{Can strong semantic transport remain stable in one-step image editing?}

To address this question, we begin by analyzing different semantic transport regimes and the factors governing their stability. Fig.~\ref{fig:motivation} reveals that, with the official GT masks from PIE-Bench~\cite{ju2024pnpinversion}, both ChordEdit and WhereEdit substantially improve non-edited-region fidelity, underscoring the importance of reliable spatial localization. Meanwhile, ChordEdit remains conservative even with accurate spatial guidance, whereas the stronger transport of WhereEdit achieves larger semantic changes but can still affect non-target regions. These observations suggest that strong semantic transport is not inherently unstable; reliable spatial localization is critical to its stability. Moreover, because a one-step update offers no later opportunity to correct misdirected transport, directional reliability is equally important. This insight naturally motivates a selective-release editing strategy: \textbf{construct a strong transport field directed toward the target semantics, but release it only at reliable spatial locations and along reliable directions.}

Guided by this principle, we propose PreciseEdit, a training-free and inversion-free framework that selectively regulates strong semantic transport based on spatial and directional reliability, thereby enabling strong target-semantic changes while preserving non-edited content with low latency. Implementing this idea, however, raises two critical challenges: \textbf{(1) How to precisely and efficiently identify where to edit?} and \textbf{(2) How to keep strong semantic transport stable?} For Challenge 1, we propose PreciseMask, a structure-aware spatial gating mechanism that combines lower-noise cross-attention semantics with self-attention structural affinity to identify reliable edit regions. PreciseMask reuses attention from the main U-Net evaluation, requiring neither external segmentation nor additional U-Net evaluations. For Challenge 2, we introduce Strong Conditional Transport (SCT), which augments source-to-target conditional contrast with anchor-relative target attraction to provide sufficient semantic modification. Cross-scale directional agreement suppresses transport directions inconsistent with a lower-noise reference, while PreciseMask provides complementary spatial regulation. Together, they enable strong transport to remain stable within a single-step update.

To effectively integrate these two modules, we propose a dual-noise-scale role decomposition strategy. The higher-noise scale provides strong semantic transport, while the lower-noise scale supplies spatial evidence and directional references. This complementary strategy couples semantic strength with transport stability within a unified one-step framework.

The main contributions can be summarized as follows:
\begin{enumerate}[leftmargin=1.8em,itemsep=4pt,topsep=4pt]
\item We propose PreciseMask, fusing lower-noise cross- and self-attention for precise, structure-aware edit localization without an external segmenter or extra U-Net evaluations.
\item We introduce Strong Conditional Transport (SCT) to strengthen target-directed transport, with cross-scale directional agreement suppressing unreliable directions.
\item We propose a dual-noise-scale strategy that leverages the complementary properties of the two noise scales to decouple transport-field construction from spatial localization, yielding leading PIE-Bench results across backbones.
\end{enumerate}
\begin{figure}[t]
 \centering\includegraphics[width=\columnwidth]{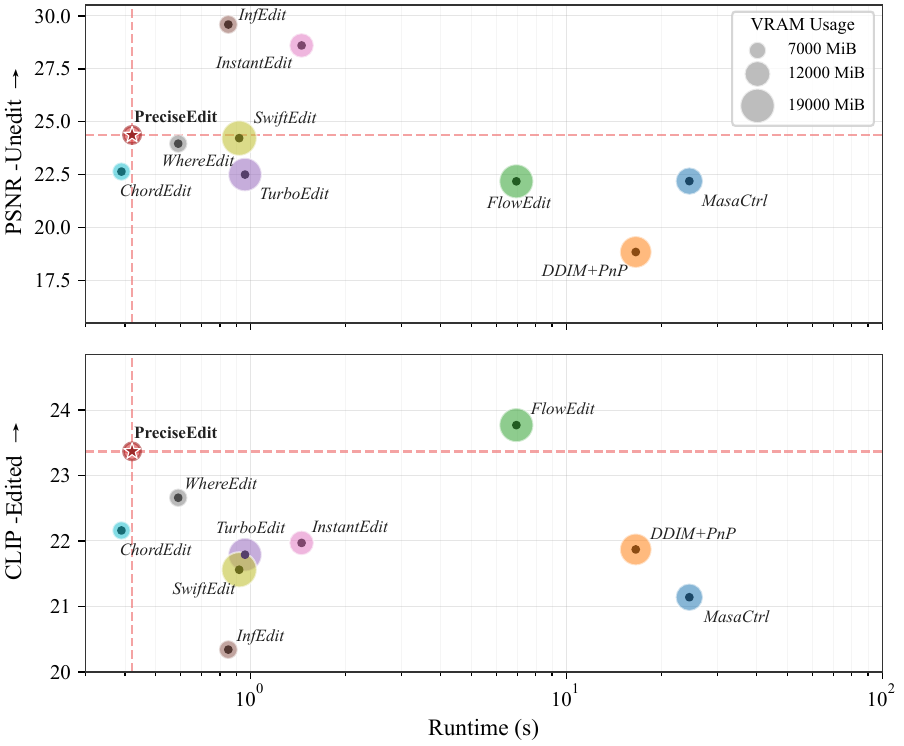}
 \caption{Quality--efficiency comparison on PIE-Bench~\cite{ju2024pnpinversion}. Each point represents an image-editing method. The horizontal axis reports runtime, while the vertical axes report background fidelity measured by PSNR-Unedit (top) and target-semantic alignment measured by CLIP-Edited (bottom)~\cite{radford2021clip}. Marker size indicates peak VRAM usage. PreciseEdit achieves a favorable balance among background preservation, semantic alignment, and inference efficiency.}
 \label{fig:efficiency}
\end{figure}
\section{Related Work}
\subsection{Text-Guided Editing with Diffusion/Flow Models}
Diffusion and latent-diffusion models established a strong foundation for text-guided visual generation and editing~\cite{hertz2023prompt2prompt,ho2020ddpm,meng2022sdedit,rombach2022ldm}. Early real-image editors commonly rely on inversion, attention control, guidance, or latent optimization~\cite{dong2023prompttuning,han2024proxedit,kawar2023imagic,mokady2023nulltext,tumanyan2023pnp,wallace2023edict}. Methods such as PnP~\cite{tumanyan2023pnp} and ProxEdit~\cite{han2024proxedit} achieve high-quality editing but require iterative inversion and multi-step sampling. Recent distilled and consistency-based generators~\cite{geng2025meanflow,nguyen2024swiftbrush,peng2026facm,song2023consistency,yin2024dmd2,yin2024dmd}, including SD-Turbo~\cite{sauer2024add}, InstaFlow~\cite{liu2024instaflow}, and SwiftBrush-v2~\cite{dao2024swiftbrushv2}, enable one- or few-step generation, but their accelerated dynamics make precise editing more challenging.
\subsection{Training-Free and Inversion-Free Editing}
To improve efficiency, recent studies avoid per-image inversion and additional model training~\cite{kulikov2025flowedit,lu2026chordedit,xu2024infedit}. FlowEdit~\cite{kulikov2025flowedit} and InfEdit~\cite{xu2024infedit} provide inversion-free editing but still rely on iterative or few-step updates. Fast editors such as TurboEdit~\cite{deutch2024turboedit}, invertible consistency distillation~\cite{starodubcev2024icd}, and SwiftEdit~\cite{nguyen2025swiftedit} reduce the sampling cost, yet compressing semantic transformation into very few updates makes fidelity and edit strength harder to balance. ChordEdit~\cite{lu2026chordedit} stabilizes one-step transport through a low-energy control field, but the resulting conservative transport can limit large semantic changes.
\subsection{Mask-Guided Controlled Image Editing}
Spatial controllability is central to precise editing. Prior work localizes modifications with explicit masks~\cite{avrahami2022blended,cao2023masactrl,couairon2023diffedit} or internal attention/features~\cite{hertz2023prompt2prompt,mao2024magedit,soni2025locatedit,tang2024locinv}. LIME~\cite{simsar2025lime} combines intermediate features and cross-attention, LayerEdit~\cite{fu2026layeredit} decomposes objects, and WhereEdit~\cite{hu2026whereedit} aggregates cross-attention to estimate editable regions. These approaches may require inversion, iterative sampling, external segmentation, or extra localization/optimization, limiting one-step efficiency. PreciseEdit instead reuses lower-noise cross- and self-attention with graph refinement, requiring no external segmenter, learnable parameters, or additional U-Net evaluation for mask generation.
\section{Method}
We propose PreciseEdit, a training-free and inversion-free framework for one-step image editing. PreciseEdit separates semantic strength from stability: the higher-noise scale provides the primary SCT, while the lower-noise scale supplies both a directional reference and attention-derived spatial support. The final update combines the primary SCT, cross-scale directional modulation, and continuous PreciseMask gating in latent space.
\subsection{Conditional Predictions in Latent Space}
Let $\mathcal{E}$ and $\mathcal{D}$ denote the frozen VAE encoder and decoder. Given a source image $I_{\ssrc}$, we encode it into latent space and use the resulting latent as the shared editing anchor:
\begin{equation} x=\mathcal{E}(I_{\ssrc}). \label{eq:anchor}\end{equation}
For a normalized scheduler timestep $t$, let $\alpha_t$ and $\sigma_t$ denote the corresponding signal and noise coefficients. We draw $n$ shared Gaussian noise samples $\{\epsilon^{(r)}\}_{r=1}^n$ and construct:
\begin{equation} z_t^{(r)}=\alpha_t x+\sigma_t\epsilon^{(r)},\quad\epsilon^{(r)}\sim\mathcal{N}(0,I).\label{eq:noising}\end{equation}
Let $P_{\ssrc}$ and $P_{\starc}$ denote the source and target prompts, and let $c\in\{\ssrc,\starc\}$ index the condition branch. To keep the formulation agnostic to the native parameterization of the one-step backbone, let $F_\theta(z,t,P_c)$ denote its native prediction and $\mathcal{C}_t(z,\cdot)$ the deterministic scheduler-specific conversion to a clean-latent estimate. We define the condition-specific clean-latent prediction as
\begin{equation}\hat\mu_c(t,x)=\frac{1}{n}\sum_{r=1}^n\mathcal{C}_t\!\left(z_t^{(r)},F_\theta(z_t^{(r)},t,P_c)\right).\label{eq:pred}\end{equation}
For an $\epsilon$-prediction diffusion backbone, for example, $\mathcal{C}_t(z,\epsilon)=(z-\sigma_t\epsilon)/\alpha_t$; flow- or velocity-parameterized backbones use the corresponding native clean-prediction conversion. The same noise samples are shared across both condition branches and noise scales, giving paired Monte Carlo estimates that reduce sampling-induced variation in conditional and cross-scale comparisons. We write the two branch predictions as $\hat\mu_{\ssrc}(t,x)$ and $\hat\mu_{\starc}(t,x)$.
\subsection{Strong Conditional Transport with Directional Agreement Modulation}
\subsubsection{Anchor-Relative Target Attraction}
Multi-step editing can correct early errors through iterative feedback, whereas one-step editing must realize sufficiently large semantic changes in a single update. We therefore seek a transport that is explicitly target-directed and adaptive to the discrepancy between the current anchor and the target-conditioned prediction.

At noise scale $t$, subtracting the source-conditioned prediction from the target-conditioned prediction gives the naive conditional residual:
\begin{equation}\Rn(t,x)=\hat\mu_{\starc}(t,x)-\hat\mu_{\ssrc}(t,x).\label{eq:naive}\end{equation}
$\Rn$ captures the source-to-target conditional contrast under the same noisy observation, but it remains a prediction-to-prediction difference and does not explicitly encode displacement from the current anchor $x$ to the target-conditioned reference. Uniformly scaling this residual changes only its magnitude. We therefore introduce an anchor-relative target attraction:
\begin{equation}a(t,x)=k[\hat\mu_{\starc}(t,x)-x].\label{eq:attraction}\end{equation}
Here $k\geq0$ balances this attraction against the conditional residual. The attraction $a(t,x)$ points directly from $x$ toward the target-conditioned clean-latent prediction, while its magnitude reflects the remaining target discrepancy. It vanishes when the two coincide.
Combining the conditional residual and anchor-relative target attraction yields Strong Conditional Transport (SCT):
\begin{equation}\Usct(t,x)=\Rn(t,x)+a(t,x).\label{eq:sct}\end{equation}
SCT therefore preserves the source-to-target conditional contrast while restoring explicit target reachability. Unlike scalar amplification of $\Rn$, its additional transport strength adapts jointly to the current anchor and target-conditioned prediction.
\subsubsection{Cross-Scale Directional Agreement Modulation}
\begin{figure*}[t]
\centering\includegraphics[width=\textwidth]{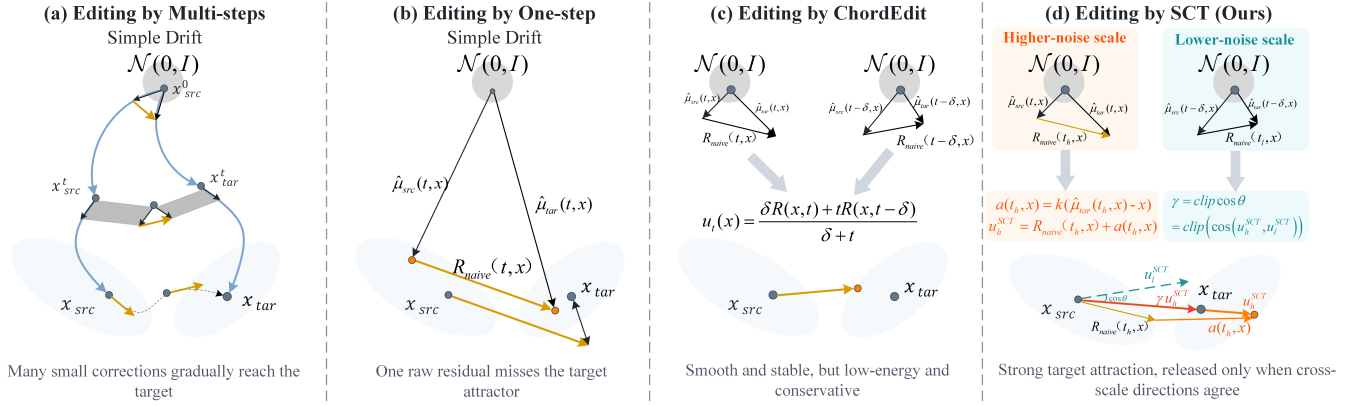}
\caption{Comparison of editing transport properties. (a) Multi-step editing: iterative updates progressively correct the trajectory toward the target. (b) Naive one-step editing: a single conditional residual may provide insufficient target attraction and miss the desired target. (c) ChordEdit~\cite{lu2026chordedit}: temporal smoothing produces a stable low-energy transport field, but can lead to conservative semantic changes. (d) SCT (Ours): anchor-relative target attraction strengthens target-directed transport, while cross-scale directional agreement suppresses unreliable directions for stable one-step editing.}
\label{fig:transport}
\end{figure*}

We construct SCT at two noise scales with complementary roles, where $t_h>t_l$. The higher-noise scale $t_h$ provides the primary SCT that enters the final latent update:
\begin{equation}u_h^{\mathrm{SCT}}=\Usct(t_h,x).\label{eq:high}\end{equation}
The lower-noise scale $t_l$ does not directly drive the final edit but rather provides a cross-scale directional reference:
\begin{equation}u_l^{\mathrm{SCT}}=\Usct(t_l,x).\label{eq:low}\end{equation}
Because transport magnitudes can differ across scales, we measure their directional agreement using cosine similarity:
\begin{equation}\gamma=\max\left\{0,\cos\left(u_h^{\mathrm{SCT}},u_l^{\mathrm{SCT}}\right)\right\}.\label{eq:agreement}\end{equation}
$\gamma$ measures agreement between the higher- and lower-noise SCT fields. Consistent directions retain stronger transport, whereas conflicting directions are suppressed. We use $\gamma u_h^{\mathrm{SCT}}$ as the directionally modulated SCT in the final composition. Figure~\ref{fig:transport} summarizes this transport construction and contrasts it with iterative correction, the naive one-step residual, and low-energy stabilization: anchor-relative attraction restores target reachability, while cross-scale agreement suppresses unreliable transport directions.
\subsection{PreciseMask}
\begin{figure*}[t]
\centering\includegraphics[width=0.95\textwidth]{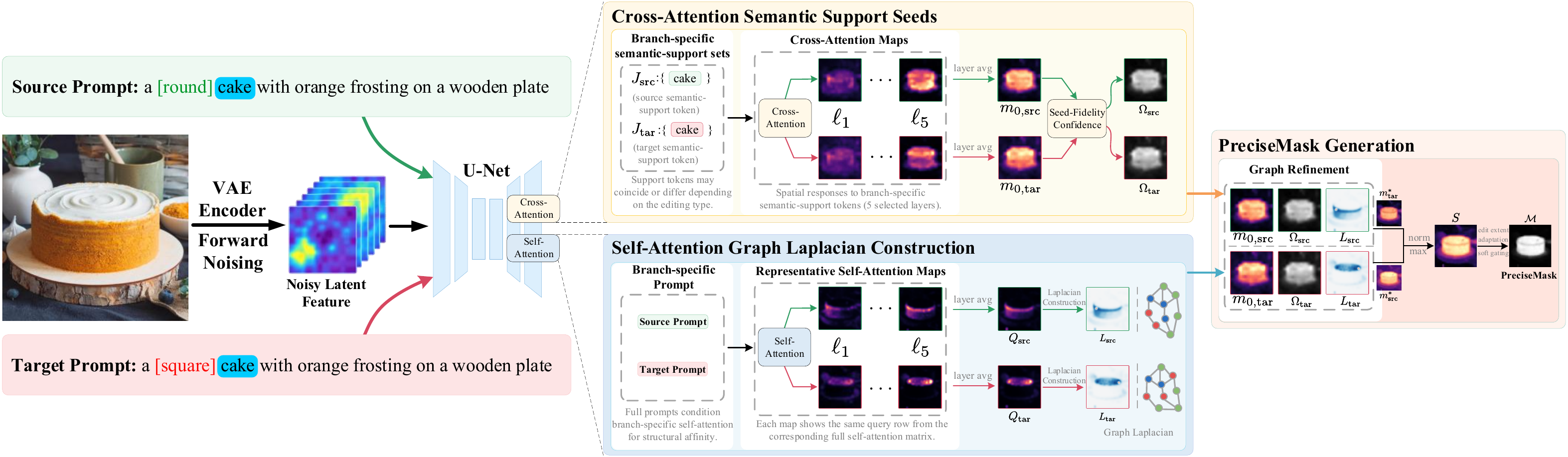}
\caption{Overview of PreciseMask. Lower-noise cross-attention produces branch-specific seeds $m_{0,c}$, which are refined with self-attention affinities to obtain $m_c^\star$. The source and target supports are combined into $S$, followed by adaptive extent estimation and soft gating to produce the final PreciseMask $M$.}\label{fig:mask}
\end{figure*}

Directional agreement determines whether the primary SCT is trustworthy, but not where it should act. We therefore use edit support to quantify spatial evidence for applying the requested latent update. PreciseMask is constructed entirely at the lower-noise scale $t_l$ by reusing cross- and self-attention from the source- and target-conditioned branches. As illustrated in Fig.~\ref{fig:mask}, branch-specific cross-attention produces semantic-support seeds, confidence-weighted self-attention graph-Laplacian refinement propagates them along coherent visual structures, and adaptive soft gating converts the refined support into PreciseMask.
\subsubsection{Cross-Attention Semantic Support Seeds}
For each condition branch $c$, we first identify its semantic support words and map them to a branch-specific set of BPE token indices, $\mathcal{J}_c$, for cross-attention aggregation. We preferentially use the blend words provided by PIE-Bench~\cite{ju2024pnpinversion} and fall back to differences between the source and target token sequences when no blend words are provided. For attribute, shape, or state edits, the two branches may share the same semantic support word, while their prompt-conditioned attention responses remain distinct. The full prompts condition the U-Net and only the selected token positions are aggregated as spatial evidence.

Let $A_{c,h}^{\ell}(:,i)$ denote the standard cross-attention probability map for token $i$ at layer $\ell$ and head $h$, $\mathcal L_x$ the selected cross-attention layers, $H_\ell$ the number of heads, and $\mathrm{Up}_\ell$ the interpolation to a common patch grid. We sum BPE-token probabilities belonging to the same support word and then average over heads and selected layers to obtain the branch-specific semantic support seed map:
\begin{equation}\label{eq:seed}
 m_{0,c}=\frac{1}{|\mathcal L_x|}\sum_{\ell\in\mathcal L_x}\mathrm{Up}_\ell\!\left(\frac1{H_\ell}\sum_{h=1}^{H_\ell}\sum_{i\in\mathcal J_c}A^{\ell}_{c,h}(:,i)\right).
\end{equation}
The seed $m_{0,c}$ provides an initial estimate of the patches associated with the semantic support of branch $c$.
\subsubsection{Self-Attention Graph Laplacian Refinement}
Cross-attention seeds can be spatially diffuse or incomplete. Inspired by LOCATEdit~\cite{soni2025locatedit}, we refine each branch seed with a self-attention graph Laplacian that propagates semantic evidence along coherent visual structure.

Let $Q_c\in\mathbb R^{N\times N}$ denote the directed patch-affinity matrix aggregated from selected self-attention layers and heads under the full prompt of branch $c$, where $Q_c(p,q)$ measures the average affinity from patch $p$ to $q$ and $N$ is the common patch-grid size. We first symmetrize the affinity:
\begin{equation}W_c=\tfrac12(Q_c+Q_c^\top).\label{eq:affinity}\end{equation}
With $Q_c^\top$ denoting the transpose and $\mathbf1$ the all-ones vector, the degree matrix is:
\begin{equation}D_c=\operatorname{diag}(W_c\mathbf1).\label{eq:degree}\end{equation}
The corresponding combinatorial graph Laplacian is:
\begin{equation}L_c=D_c-W_c.\label{eq:laplacian}\end{equation}
The Laplacian encourages strongly connected patches to receive similar support scores. To prevent propagation from overriding reliable cross-attention seeds, we define a position-wise seed-fidelity confidence:
\begin{equation}\omega_c(p)=\operatorname{sigmoid}\!\left(\kappa[m_{0,c}(p)-b_\omega]\right).\label{eq:confidence}\end{equation}
Here $\operatorname{sigmoid}(\cdot)$ is the logistic function, while $\kappa$ and $b_\omega$ control its slope and center.

We place the squared confidences on the diagonal of $\Omega_c$, which weights how strongly each patch should retain its original seed:
\begin{equation}\Omega_c=\operatorname{diag}(\omega_c(1)^2,\dots,\omega_c(N)^2).\label{eq:omega}\end{equation}
Let $m\in\mathbb R^N$ be the optimization variable for the patch-wise edit-support vector. Using $m_{0,c}$ as the seed reference, we jointly optimize confidence-weighted seed fidelity and graph smoothness:
\begin{equation}\label{eq:graphopt}\begin{aligned}m_c^\star=\arg\min_{m\in\mathbb R^N}\; &(m-m_{0,c})^\top\Omega_c(m-m_{0,c})\\&+\lambda m^\top L_cm.\end{aligned}\end{equation}
The data term anchors reliable seed responses, while the Laplacian term propagates support across high-affinity patches and suppresses isolated activations. The coefficient $\lambda$ balances seed fidelity and structural smoothness, and the optimizer $m_c^\star$ is the graph-refined branch support.
We compute the unique minimizer with the equivalent linear system:
\begin{equation}(\Omega_c+\lambda L_c)m_c^\star=\Omega_cm_{0,c}.\label{eq:linear}\end{equation}
We normalize the two graph-refined branch supports and retain the stronger response at each patch to form the structure-refined edit-support map:
\begin{equation}S(p)=\max\left\{[\operatorname{Norm}(m_{\ssrc}^\star)]_p,[\operatorname{Norm}(m_{\starc}^\star)]_p\right\}.\label{eq:support}\end{equation}
$\operatorname{Norm}(\cdot)$ denotes per-map min--max normalization to $[0,1]$. $S$ integrates source- and target-conditioned semantic support while retaining spatial coherence through self-attention affinities. For object replacement, the two branches naturally provide removal and insertion evidence, respectively.
\subsubsection{Adaptive Edit Extent}
A fixed threshold or top percentage imposes a similar spatial budget on heterogeneous edits. To keep local edits compact while allowing diffuse changes such as weather or style to expand, we adapt the edit extent to the concentration of $S$.
For numerical stability, let $\epsilon_e>0$. We first compute a normalized participation ratio:
\begin{equation}e(S)=\frac{\left(\sum_{p=1}^NS(p)\right)^2}{N\sum_{p=1}^NS(p)^2+\epsilon_e}.\label{eq:participation}\end{equation}
Smaller $e(S)$ indicates concentrated support and larger values indicate diffuse support. We map it to a target cumulative support mass $\rho(S)$, with $\rho_0$, $e_0$, and $a_\rho$ controlling the mapping and $\rho_{\min}$, $\rho_{\max}$ bounding its range:
\begin{equation}\rho(S)=\operatorname{clip}\big(\rho_0+a_\rho[e(S)-e_0],\rho_{\min},\rho_{\max}\big).\label{eq:rho}\end{equation}
We sort the support scores in descending order and select the smallest cumulative set whose normalized mass reaches $\rho(S)$; the lowest score in this set defines the adaptive threshold $\tau$.
Let $T>0$ denote the gating temperature and define $g_{\min}=\operatorname{sigmoid}(-\tau/T)$ and $g_{\max}=\operatorname{sigmoid}((1-\tau)/T)$. Endpoint normalization yields a continuous PreciseMask $M\in[0,1]^N$:
\begin{equation}M(p)=\frac{\operatorname{sigmoid}((S(p)-\tau)/T)-g_{\min}}{g_{\max}-g_{\min}}.\label{eq:maskgate}\end{equation}
Thus, PreciseMask is a position-wise continuous spatial gate.
\subsection{Strong-Yet-Stable Latent Composition}
At this stage, SCT provides semantic strength, $\gamma$ provides directional agreement, and PreciseMask $M$ provides position-wise spatial gating. After interpolating $M$ to the latent resolution and sharing it across channels, we form the intermediate latent:
\begin{equation}\tilde x=x+\eta M\odot\left(\gamma u_h^{\mathrm{SCT}}\right).\label{eq:middle}\end{equation}
Here $\eta>0$ is the transport step size.
Following ChordEdit~\cite{lu2026chordedit}, we then apply target-conditioned proximal refinement to the intermediate latent $\tilde x$ at refinement timestep $t_r$, yielding $x_{\mathrm{prox}}$. The final edited latent is composited as:
\begin{equation}x_{\mathrm{out}}=(1-M)\odot x+M\odot x_{\mathrm{prox}}.\label{eq:out}\end{equation}
Define the target-conditioned refinement residual as $r_{\mathrm{prox}}=x_{\mathrm{prox}}-\tilde x$. Expanding Eq.~\eqref{eq:out} gives:
\begin{equation}x_{\mathrm{out}}=x+\eta\gamma M^2\odot u_h^{\mathrm{SCT}}+M\odot r_{\mathrm{prox}}.\label{eq:expand}\end{equation}
Here $M^2=M\odot M$ denotes element-wise squaring. The primary SCT is therefore attenuated quadratically by the soft gate, whereas the refinement residual is weighted linearly by $M$. Strong transport is preserved where $M\approx1$ and decays faster near uncertain boundaries, realizing selective release according to directional and spatial reliability. The final image is decoded by the frozen VAE decoder as $I_{\mathrm{out}}=\mathcal D(x_{\mathrm{out}})$.
\begin{table*}[t]
\caption{Quantitative comparison on PIE-Bench~\cite{ju2024pnpinversion}.
T-free/I-free denote training-/inversion-free.
The best/second/third results in each numeric column are highlighted with
\colorbox{ranksecond}{\strut orange}/
\colorbox{rankbest}{\strut yellow}/
\colorbox{rankthird}{\strut blue} backgrounds, respectively.
The GT-guided oracle row is excluded from ranking.}
\label{tab:main}
\centering
\scriptsize
\setlength{\tabcolsep}{3.1pt}
\renewcommand{\arraystretch}{1.12}

\resizebox{\textwidth}{!}{%
\begin{tabular}{@{}ll!{\vrule width 0.4pt}ccc!{\vrule width 0.4pt}cc!{\vrule width 0.4pt}cc!{\vrule width 0.4pt}ccc@{}}
\toprule

\multirow{2}{*}{Type}
& \multirow{2}{*}{Method}
& \multicolumn{3}{!{\vrule width 0.4pt}c}{Consistency}
& \multicolumn{2}{!{\vrule width 0.4pt}c}{CLIP Semantics}
& \multicolumn{2}{!{\vrule width 0.4pt}c}{Properties}
& \multicolumn{3}{!{\vrule width 0.4pt}c@{}}{Efficiency}
\\

\cmidrule(lr){3-5}
\cmidrule(lr){6-7}
\cmidrule(lr){8-9}
\cmidrule(lr){10-12}

&
& PSNR$\uparrow$
& MSE$_{10^3}\downarrow$
& LPIPS$_{10^3}\downarrow$
& Whole$\uparrow$
& Edited$\uparrow$
& T-free
& I-free
& Step$\downarrow$
& Runtime(s)$\downarrow$
& VRAM(MiB)$\downarrow$
\\

\midrule

\multirow{3}{*}{\shortstack{Multi-step\\($\geq30$ steps)}}
& DDIM + MasaCtrl~\cite{song2021ddim,cao2023masactrl}
& 22.19 & 8.64 & 105.44
& 24.02 & 21.14
& \cmark & \xmark
& 50 & 24.49 & 13016
\\

& DDIM + PnP~\cite{tumanyan2023pnp}
& 18.84 & 18.42 & 183.97
& 24.97 & 21.87
& \cmark & \xmark
& 50 & 16.57 & 16494
\\

& FlowEdit (SD3)~\cite{kulikov2025flowedit}
& 22.18 & 8.74 & 105.20
& \secondcell{26.81}
& \secondcell{23.77}
& \cmark & \cmark
& 33 & 6.94 & 18536
\\

\midrule

\multirow{3}{*}{\shortstack{Few-step\\(4 steps)}}
& TurboEdit (SDXL-Turbo)~\cite{deutch2024turboedit}
& 22.50 & 9.30 & 105.57
& 24.68 & 21.79
& \cmark & \cmark
& 4 & 0.96 & 18004
\\

& InfEdit (SD1.4)~\cite{xu2024infedit}
& \secondcell{29.59}
& \secondcell{2.26}
& \secondcell{37.10}
& 23.04 & 20.34
& \cmark & \cmark
& 4 & 0.85 & 7620
\\

& InstantEdit (PeRFlow-SD1.5)~\cite{gong2025instantedit}
& \bestcell{28.60}
& \bestcell{2.80}
& \bestcell{44.94}
& 25.09 & 21.97
& \cmark & \xmark
& 4 & 1.45 & 10994
\\

\midrule

\multirow{4}{*}{One-step}

& SwiftEdit (SwiftBrush-v2)~\cite{nguyen2025swiftedit}
& 24.22
& \thirdcell{5.41}
& \thirdcell{76.39}
& 24.48
& 21.56
& \xmark
& \xmark
& \secondcell{1}
& 0.92
& 19628
\\

& ChordEdit (SD-Turbo)~\cite{lu2026chordedit}
& 22.64
& 8.00
& 118.50
& 24.82
& 22.16
& \cmark
& \cmark
& \secondcell{1}
& \secondcell{0.39}
& \bestcell{7418}
\\

& WhereEdit~\cite{hu2026whereedit}
& 23.96
& 8.05
& 108.91
& \thirdcell{25.21}
& \thirdcell{22.66}
& \cmark
& \cmark
& \secondcell{1}
& \thirdcell{0.59}
& \thirdcell{7436}
\\

& \textbf{PreciseEdit (ours)}
& \thirdcell{24.37}
& 8.33
& 98.66
& \bestcell{25.45}
& \bestcell{23.36}
& \cmark
& \cmark
& \secondcell{1}
& \bestcell{0.42}
& \secondcell{6954}
\\

\midrule

\rowcolor{oraclegray}
& \textbf{PreciseEdit (w/ GT, ours)}
& 27.34
& 2.65
& 43.20
& 25.77
& 23.90
& \cmark
& \cmark
& 1
& 0.38
& 6936
\\

\bottomrule
\end{tabular}%
}
\vspace{5pt}

\centering\includegraphics[width=0.89\textwidth]{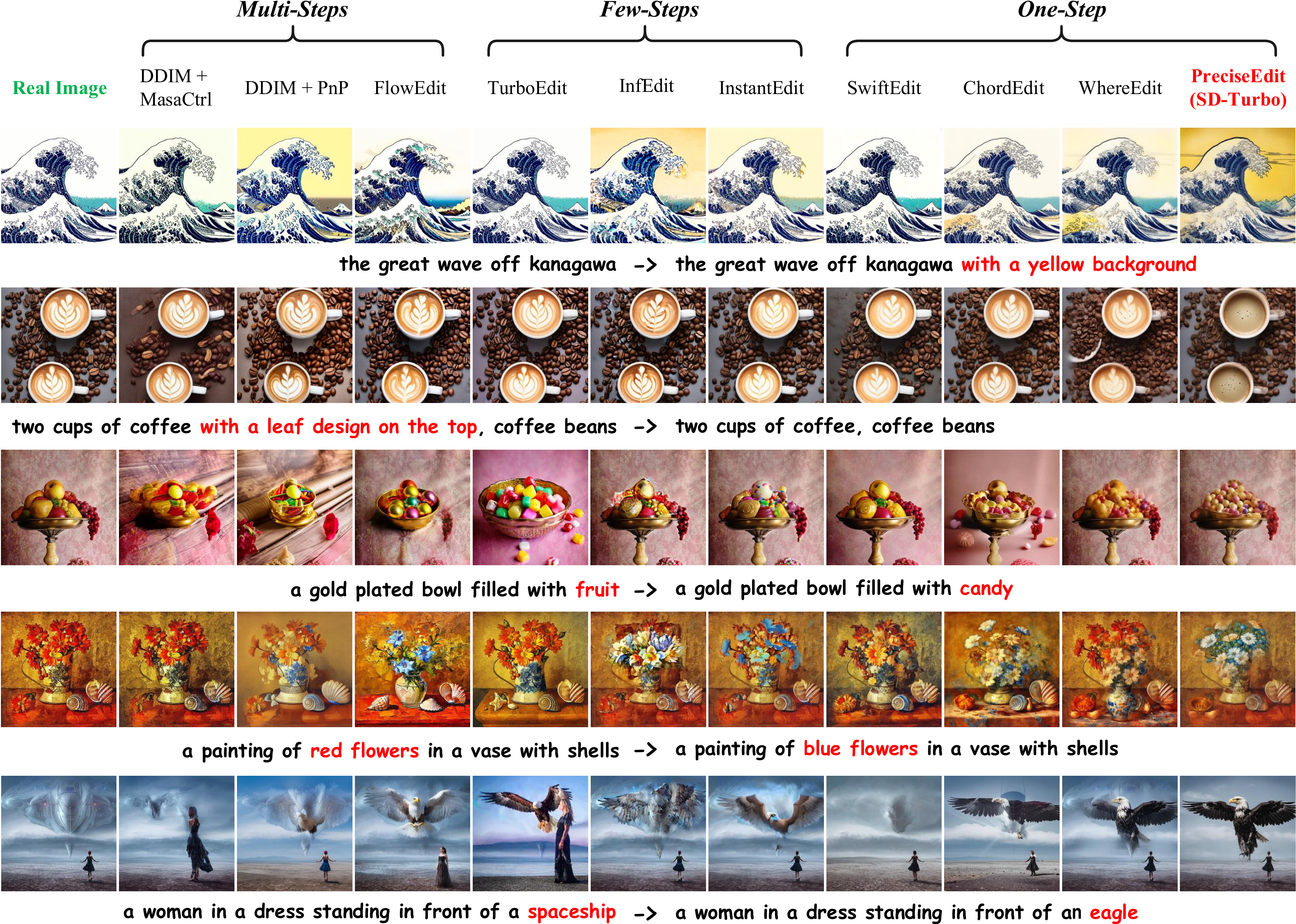}
\captionof{figure}{Comparison of edited results. Real images are in the first column. Prompts are noted under each row.}\label{fig:comparisons}

\end{table*}
\section{Experiments}
\subsection{Experimental Setup}
\noindent\textbf{Dataset and evaluation metrics.} We conduct our empirical evaluation on PIE-Bench~\cite{ju2024pnpinversion}, a standard benchmark for instruction-based image editing on $512\times512$ images. The benchmark contains 700 samples distributed across 10 distinct editing categories, with each sample providing a source image, source and target textual prompts, blend word, and a GT mask specifying the editing region. We assess performance along two complementary axes: background fidelity and semantic alignment. Background fidelity is measured using PSNR, MSE, and LPIPS on the non-edited regions, while semantic alignment is evaluated using CLIP-Whole and CLIP-Edited~\cite{radford2021clip} for the entire image and edited region, respectively.

\noindent\textbf{Implementation details.} Code will be made publicly available upon acceptance. Our default backbone is SD-Turbo~\cite{sauer2024add}, evaluated in FP32 on a single NVIDIA GeForce RTX 3090 GPU with 24 GB of memory, using one shared Gaussian noise sample ($n=1$). We set the higher-noise scale to $t_h=0.90$, the lower-noise scale to $t_l=0.35$, and the target-conditioned refinement scale to $t_r=0.30$. The SCT attraction coefficient is $k=2.0$, and the latent step size is 1.0. One batched U-Net evaluation processes the source- and target-conditioned branches at both noise scales, followed by one target-only evaluation for proximal refinement.

\begin{figure*}[t]
\centering\includegraphics[width=.94\textwidth]{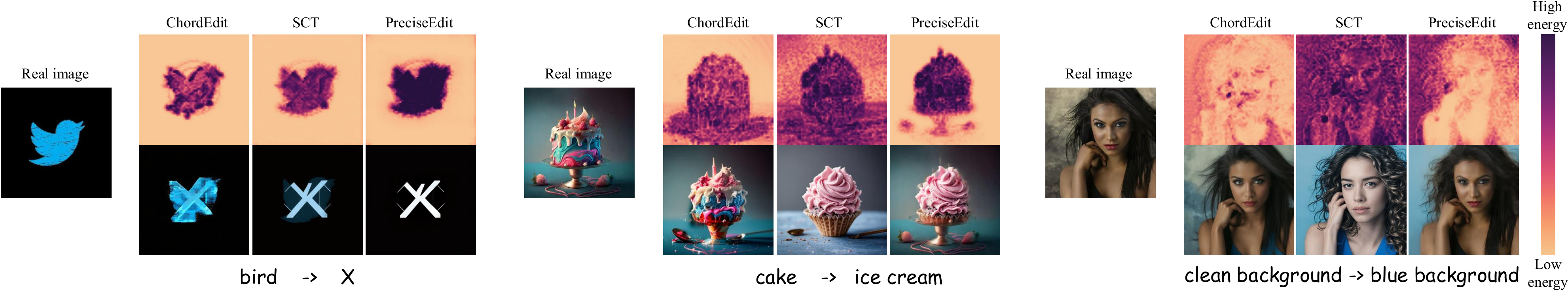}
\caption{Transport-energy localization and editing outcomes. ChordEdit~\cite{lu2026chordedit} yields a comparatively weak field, whereas ungated SCT spreads strong transport beyond the intended region. PreciseMask preserves high transport energy over reliable edit support while suppressing it elsewhere.}\label{fig:energy}
\end{figure*}
\subsection{Comparison with Prior Methods}
\noindent\textbf{Quantitative Results.} Table~\ref{tab:main} compares PreciseEdit with multi-step, few-step, and one-step editors on PIE-Bench. Overall, our training-free, inversion-free method combines leading semantic fidelity with outstanding efficiency, requiring only 0.42 s and the least VRAM (6,954 MiB) among non-oracle methods. Compared with multi-step methods, it achieves higher PSNR-Unedit than all baselines while running $16.5\times$ faster than FlowEdit and $58.3\times$ faster than DDIM+MasaCtrl. Against few-step editors, PreciseEdit delivers the highest CLIP-Edited, at least $2.0\times$ faster inference, and the lowest VRAM. Within the one-step category, it ranks first in both PSNR-Unedit (24.37 dB) and CLIP-Edited (23.36), improving over ChordEdit~\cite{lu2026chordedit} by 1.73 dB/1.20 and WhereEdit~\cite{hu2026whereedit} by 0.41 dB/0.70. The GT-guided variant attains 27.34 dB/23.90, highlighting further potential from precise localization.

\noindent\textbf{Qualitative Results.} Figure~\ref{fig:comparisons} visually reinforces the quantitative gains. Compared with multi-step methods, PreciseEdit produces cleaner edits while preserving source structures. Against few-step editors, it achieves faithful target transformations with strong visual coherence. Among one-step methods, it achieves more complete semantic changes than methods such as ChordEdit while better preserving unrelated details. Across background recoloring, object replacement, and attribute editing, PreciseEdit delivers accurate target semantics while retaining source composition.
\subsection{Ablation Studies}
Unless stated otherwise, all ablations use the default PIE-Bench and SD-Turbo setting.

\begin{figure*}[t]
\centering\includegraphics[width=.98\textwidth]{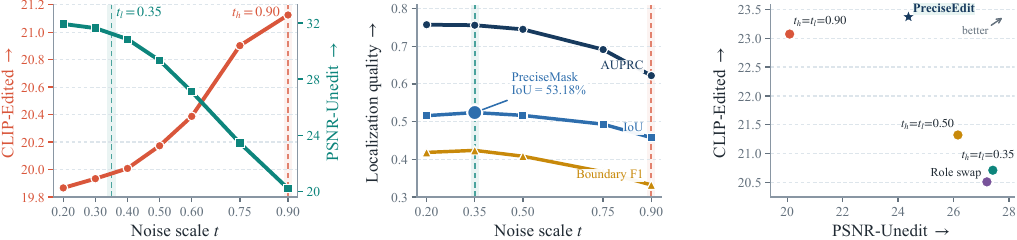}
\caption{Analysis of dual-noise-scale role specialization. Left: Transport-scale analysis. An isolated conditional-residual probe shows that increasing the noise scale strengthens semantic editing but reduces source preservation, indicating greater semantic reach at higher noise. Middle: Localization-scale analysis. Varying only the attention scale for PreciseMask shows that lower-noise attention provides more reliable spatial support; $t_l=0.35$ achieves the best IoU and Boundary F1 while retaining near-maximum AUPRC. Right: Role-allocation analysis. With full PreciseEdit, assigning $t_h=0.90$ to semantic transport and $t_l=0.35$ to spatial guidance yields a substantially better fidelity--editability balance than shared-scale or role-swapped alternatives.}\label{fig:scales}
\end{figure*}

\noindent\textbf{Analysis of strong-transport localization.} Figure~\ref{fig:energy} shows that ChordEdit remains globally weak, whereas ungated SCT spreads strong energy into source-preserving regions. PreciseMask confines high transport energy to reliable edit support, retaining semantic strength while reducing background and structural distortion.
\begin{table}[t]
\caption{Mask-generator comparison under the same SCT field. PreciseMask improves localization and editing quality while reducing runtime and memory relative to LIME~\cite{simsar2025lime} and WhereEdit~\cite{hu2026whereedit}.}
\label{tab:maskquality}
\centering
\footnotesize
\setlength{\tabcolsep}{2pt}

\resizebox{\columnwidth}{!}{\begin{tabular*}{\columnwidth}{@{\extracolsep{\fill}}>{\raggedright\arraybackslash}p{0.33\columnwidth}ccccc@{}}
\toprule
Method & PSNR$\uparrow$ & CLIP-Edited$\uparrow$ & RunTime(s)$\downarrow$ & VRAM$\downarrow$ & IoU(\%)$\uparrow$\\
\midrule
SCT + LIME Mask~\cite{simsar2025lime}& 24.18 & 21.57 & 0.74 & 7450 & 16.59\\
SCT + WhereEdit Mask~\cite{hu2026whereedit}& 23.09 & 22.89 & 0.59 & 7436 & 44.19\\
\textbf{SCT + PreciseMask} & \textbf{24.37} & \textbf{23.36} & \textbf{0.42} & \textbf{6954} & \textbf{53.18}\\
\bottomrule
\end{tabular*}}
\end{table}
\noindent\textbf{Analysis of mask quality and efficiency.} Table~\ref{tab:maskquality} fixes SCT and changes only the mask generator. PreciseMask reaches 53.18\% IoU, versus 44.19\% for WhereEdit and 16.59\% for LIME, while attaining the best CLIP-Edited, runtime, and VRAM. Thus, its localization gain requires no extra U-Net evaluation, whereas WhereEdit~\cite{hu2026whereedit} performs a separate pass for mask generation.
\begin{table}[t]
\captionsetup{skip=4pt}
\caption{Cross-backbone generalization. PreciseEdit consistently improves both PSNR-Unedit and CLIP-Edited over ChordEdit~\cite{lu2026chordedit} on three fast T2I backbones.}
\label{tab:backbones}
\centering
\footnotesize
\setlength{\tabcolsep}{3.6pt}
\renewcommand{\arraystretch}{0.92}
\resizebox{\columnwidth}{!}{\begin{tabular*}{\columnwidth}{@{\extracolsep{\fill}}lcccc@{}}
\toprule
\multirow{2}{*}{T2I Backbone} & \multicolumn{2}{c}{PSNR$\uparrow$} & \multicolumn{2}{c}{CLIP-Edited$\uparrow$}\\
\cmidrule(r{0.35em}){2-3}\cmidrule(l{0.35em}){4-5}
& ChordEdit & Ours & ChordEdit & Ours\\
\midrule
InstaFlow~\cite{liu2024instaflow} & 21.47 & 23.95 & 20.43 & 21.66\\
SwiftBrush-v2~\cite{dao2024swiftbrushv2} & 22.89 & 23.76 & 21.23 & 22.85\\
SD-Turbo~\cite{sauer2024add} & 22.64 & 24.37 & 22.16 & 23.36\\
\bottomrule
\end{tabular*}}
\end{table}
\noindent\textbf{Analysis of dual-noise-scale role specialization.} Figure~\ref{fig:scales} validates the complementary roles of the two noise scales through three controlled analyses. In the transport-scale analysis (left), we vary only the noise scale of the conditional residual; higher noise improves semantic editability at the cost of source preservation. In the localization-scale analysis (middle), we vary only the attention extraction scale; lower-noise attention provides more reliable spatial support, with $t_l=0.35$ achieving the best IoU (53.18\%) and Boundary F1 (42.35\%). In the role-allocation analysis (right), full PreciseEdit with $t_h=0.90$ for semantic transport and $t_l=0.35$ for spatial guidance yields a better fidelity--editability balance than shared-scale or role-swapped alternatives. Together, the two scales are complementary rather than interchangeable.

\noindent\textbf{Analysis of cross-backbone generalization.} Table~\ref{tab:backbones} applies the same training-free formulation to InstaFlow~\cite{liu2024instaflow}, SwiftBrush-v2~\cite{dao2024swiftbrushv2}, and SD-Turbo~\cite{sauer2024add}, comparing against ChordEdit on the same backbone. PreciseEdit improves PSNR-Unedit and CLIP-Edited by 2.48/1.23, 0.87/1.62, and 1.73/1.20, respectively. Consistent gains across flow- and diffusion-based fast generators support the model-agnostic design.
\noindent\textbf{Analysis of component contributions.} Table~\ref{tab:components} evaluates three ablation blocks. Block A highlights the individual strengths and complementarity of SCT and PreciseMask. Ungated SCT raises CLIP-Edited from 22.57 to 24.25, while PreciseMask improves naive-transport PSNR-Unedit by 7.65 dB. Together, they recover 9.81 dB over ungated SCT, reaching 24.37 dB PSNR-Unedit and 23.36 CLIP-Edited. Block B shows that target attraction boosts CLIP-Edited by 3.25 over $k=0$, while directional agreement gains 3.08 dB PSNR-Unedit and reduces LPIPS from 0.1390 to 0.0987. Block C shows that graph refinement ($\Omega=I$) improves IoU and Boundary F1 by 1.19 and 1.66 points. Under adaptive extent, confidence weighting further gains 1.23 IoU, 3.43 Boundary F1, and 0.32 CLIP-Edited points. Replacing adaptive extent with fixed mass ($\rho=0.75$) reduces IoU and Boundary F1 by 5.44 and 5.93 points.
\begin{table}[t]
\centering
\caption{\textbf{Component ablation on PIE-Bench.}
Block A studies SCT--PreciseMask interaction; Block B ablates SCT
with PreciseMask fixed. Block C uses controlled pairs for graph
refinement (rows 1--2), confidence weighting (rows 2 and 4), and
adaptive extent (rows 3--4), with SCT fixed. Best results per block
are bold; ``--'' denotes unevaluated metrics.}
\label{tab:components}

\footnotesize
\setlength{\tabcolsep}{3.0pt}
\renewcommand{\arraystretch}{1.08}
\resizebox{\columnwidth}{!}{
\begin{tabular}{lccccc}
\toprule
Variant
& PSNR$\uparrow$
& LPIPS$\downarrow$
& CLIP-Edited$\uparrow$
& IoU(\%)$\uparrow$
& BF1(\%)$\uparrow$ \\
\midrule

\multicolumn{6}{l}{
\textbf{A. Major Components (SCT $\times$ PreciseMask)}} \\

Naive Transport
& 20.13 & -- & 22.57 & -- & -- \\

Naive Transport + PreciseMask
& \textbf{27.78} & -- & 21.63 & -- & -- \\

Ungated SCT
& 14.56 & -- & \textbf{24.25} & -- & -- \\

Full PreciseEdit
& 24.37 & 0.0987 & 23.36 & 53.18 & 42.35 \\

\midrule

\multicolumn{6}{l}{
\textbf{B. SCT (full PreciseMask)}} \\

w/o Target Attraction ($k=0$)
& \textbf{29.40}
& \textbf{0.0502}
& 20.11
& -- & -- \\

w/o Directional Agreement ($\gamma=1$)
& 21.29
& 0.1390
& 23.21
& -- & -- \\

Full SCT
& 24.37
& 0.0987
& \textbf{23.36}
& -- & -- \\

\midrule

\multicolumn{6}{l}{
\textbf{C. PreciseMask (full SCT)}} \\

Cross-Attn. Seeds (adaptive)
& \textbf{25.25}
& \textbf{0.0839}
& 22.84
& 50.76
& 37.26 \\

Graph, $\Omega=I$ (adaptive)
& 24.94
& 0.0879
& 23.04
& 51.95
& 38.92 \\

Graph, $\Omega_c$ (fixed $\rho=0.75$)
& 24.50
& 0.0920
& 23.12
& 47.74
& 36.42 \\

\makebox[0pt][l]{Full PreciseMask (adaptive)}%
\phantom{Full PreciseMask (Adaptive Edit Extent)}
& 24.37
& 0.0987
& \textbf{23.36}
& \textbf{53.18}
& \textbf{42.35} \\

\bottomrule
\end{tabular}
}
\end{table}
\noindent\textbf{Analysis of mask transferability.} Table~\ref{tab:transfer} shows that PreciseMask raises PSNR-Unedit for all four transport rules; on WhereEdit, it improves PSNR-Unedit/CLIP-Edited from 23.96/22.66 to \textbf{25.36/22.95}, confirming transferability beyond SCT. GT-guided rows indicate further localization headroom.
\begin{table}[t]
\caption{Effect and transferability of spatial-mask strategies. We insert no mask, the official GT mask, and PreciseMask into the naive, ChordEdit~\cite{lu2026chordedit}, WhereEdit~\cite{hu2026whereedit}, and PreciseEdit transport rules. GT-guided rows provide an oracle upper bound on localization quality.}
\label{tab:transfer}
\centering
\scriptsize
\setlength{\tabcolsep}{3.2pt}
\renewcommand{\arraystretch}{1.27}
\resizebox{\columnwidth}{!}{\begin{tabular*}{\columnwidth}{@{\extracolsep{\fill}}llcc@{}}
\toprule
Method & Mask Strategy & PSNR$\uparrow$ & CLIP-Edited$\uparrow$\\
\midrule
\multirow{3}{*}{Naive}
& w/o Mask & 20.13 & 22.57\\
& w/ GT Mask & 31.21 & 22.23\\
& w/ PreciseMask & 27.78 & 21.63\\
\midrule
\multirow{3}{*}{ChordEdit~\cite{lu2026chordedit}}
& w/o Mask & 22.64 & 22.16\\
& w/ GT Mask & 31.38 & 21.77\\
& w/ PreciseMask & 28.55 & 21.30\\
\midrule
\multirow{4}{*}{WhereEdit~\cite{hu2026whereedit}}
& w/o Mask & 16.34 & 24.36\\
& w/ GT Mask & 30.21 & 23.63\\
& w/ WhereEdit Mask (original) & 23.96 & 22.66\\
& w/ PreciseMask (ours) & 25.36 & 22.95\\
\midrule
\multirow{3}{*}{PreciseEdit (ours)}
& w/o Mask (SCT) & 14.56 & 24.25\\
& w/ GT Mask & 27.34 & 23.90\\
& \textbf{w/ PreciseMask} & \textbf{24.37} & \textbf{23.36}\\
\bottomrule
\end{tabular*}}
\end{table}

\section{Conclusion}
In this work, we revisited one-step image editing through the tension between semantic strength and transport reliability. Instead of globally suppressing strong transport, we proposed PreciseEdit, a training-free and inversion-free framework that selectively trusts it according to directional and spatial reliability. Our dual-noise-scale design uses a higher-noise scale to construct Strong Conditional Transport, while a lower-noise scale provides directional calibration and structured spatial support through PreciseMask. By combining cross-attention semantic seeds, self-attention graph-Laplacian refinement, and adaptive soft gating, PreciseEdit localizes strong semantic changes while preserving irrelevant regions. Extensive experiments on PIE-Bench~\cite{ju2024pnpinversion} demonstrate that PreciseEdit achieves a favorable fidelity--editability Pareto balance while retaining low latency and memory usage. Its consistent gains across multiple one-step backbones further support the generality of the proposed formulation. We hope that the perspective of separating semantic strength from reliability will provide a useful foundation for efficient, precise, and controllable image editing.

\clearpage
\renewcommand{\bibfont}{\fontsize{8.45}{10.1}\selectfont}
\bibliographystyle{unsrt}
\bibliography{preciseedit_refs_arxiv}
\end{document}